\documentclass{article}
\usepackage{stywhispers,amsmath,epsfig}
\usepackage{color}
\usepackage[numbers,sort&compress]{natbib}
\usepackage{balance}

\title{Person Re-identification with Hyperspectral Multi-Camera Systems --- A Pilot Study}
\name{Saurabh Prasad$^{1}$, Tanu Priya$^{1}$, Minshan Cui$^{1}$, Shishir Shah$^{2}$ \thanks{Corresponding Author:~Saurabh Prasad; Email: saurabh.prasad@ieee.org. This work is funded in part by National Aeronautics and Space Administration and Army Research Office.}}
\address{University of Houston}
\address{
$^{1}$\,Department of Electrical and Computer Engineering \\ 
University of Houston 
\\
$^{2}$\,Department of Computer Science \\
University of Houston
\\[0.5em]
}
\begin{document}
%
\maketitle
\begin{abstract}
Person re-identification in a multi-camera environment is an important part of modern surveillance systems. Person re-identification from color images has been the focus of much active research, due to the numerous challenges posed with such analysis tasks, such as variations in illumination, pose and viewpoints. In this paper, we suggest that hyperspectral imagery has the potential to provide unique information that is expected to be beneficial for the re-identification task. Specifically, we assert that by accurately characterizing the unique spectral signature for each person's skin, hyperspectral imagery can provide very useful descriptors (e.g. spectral signatures from skin pixels) for re-identification. 
Towards this end, we acquired proof-of-concept hyperspectral re-identification data under challenging (practical) conditions from 15 people. Our results indicate that hyperspectral data result in a substantially enhanced re-identification performance compared to color (RGB) images, when using spectral signatures over skin as the feature descriptor.
\end{abstract}
\section{Introduction}
\label{sec:intro}

Hyperspectral imaging systems have become increasingly popular for a variety of applications, including remote sensing and biomedical analytics. With their dense, contiguous and narrow-band spectral sampling in visible through short-wave infrared regions of the electromagnetic spectrum, they provide rich spectral characterization of the objects that are dominant in the pixels of the hyperspectral image. The capability of hyperspectral data to accurately capture material-specific properties makes them an attractive choice for characterizing vegetation mapping invasive and endangered vegetation, (e.g., for ecological and precision agriculture applications), detection and characterization of physiological conditions and other related biomedical applications \cite{vyas2013estimating,li2012compressive,chen2011denoising,NG2011,prasad2014SMoG,PLFB2012,KPB2010Derivatives,prasad2009information,YP2015,PC2013Asilomar_Sparse,li2011multi,di2011active,PB2008a}. In addition to such applications where hyperspectral data has seen a widespread popularity and acceptance, we suggest that hyperspectral data is also particularly suited for enhanced computer vision applications as they relate to scene understanding, biometrics, and person re-dentification, by virtue of their robust characterization of material properties. 

Person re-identification -- the task of recognizing a person separated in location and time, has emerged as an important application of multi-camera surveillance systems. Despite algorithmic advances \cite{gheissari2006person,farenzena2010person,zheng2011person} built upon traditional computer vision imaging systems, person reidentification remains a difficult problem due to various challenges, including variations in illumination conditions, pose and viewpoints. Advances in this area include pose, viewpoint and illumination invariant feature extraction \cite{gheissari2006person,farenzena2010person}, as well as the design of appropriate similarity metrics. A majority of these methods take into account global appearance of a person, such as a weighted color histogram \cite{farenzena2010person}. Local spatial information is often extracted by analyzing cropped regions around the face \cite{dantcheva2011frontal}. 

In this paper, we assert that hyperspectral imagery is potentially very beneficial for the task of person re-identification in a multi-camera surveillance scenario --- specifically, spectral content can serve as a powerful descriptor that can be used by itself or in conjunction with classical spatial and statistical features derived for re-identification. The spectral reflectance \emph{signature} will demonstrate variability across materials (e.g. clothing) and across skin between different people. While it is expected that traditional computer vision imaging systems can characterize differences in clothing and other traditional descriptors, which may be highly separable between people in the visible wavelength regime, subtle differences due to individual skin physiology can be better characterized by hyperspectral imagery. 
With this in mind, we developed a pilot study and acquired hyperspectral imagery from fifteen subjects in the visible and very near infrared region of the electromagnetic spectrum at two different locations and at different times of the day (morning versus afternoon). We utilized a simple distance metric appropriate for hyperspectral data and reported re-identification performance using spectral features derived from superpixel patches over the skin, to demonstrate the efficacy of such data for person reidentification. This was compared with reidentification undertaken with color imagery.

The outline of this paper is as follows. In section 2, we provide details of the data acquisition, and describe the data that was acquired. In section 3, we describe the approach utilized to setup and quantify re-identification performance with this data. In section 4, we provide experimental results for the re-identification task, and in section 5, we provide concluding remarks.

\section{Hyperspectral Data Acquisition for Person Reidenitifcation}
\label{sec:data}

With the goal of demonstrating the efficacy of hyperspectral data for person re-identification, we acquired hyperspectral data from $15$ people under a traditional multi-camera surveillance scenario at two different locations and two different times of the day (morning and afternoon). The hyperspectral images were acquired using a Headwall Photonics$^\text{TM}$ hyperspectral imager --- the spatial size of each image was $1004\times400$ and each pixel had $325$ contiguous spectral bands spanning the visible and near-infrared spectrum from $400nm-1000nm$ uniformly, with a full-width half-maximum (FWHM) bandwidth of $1.8nm$.  In order to evaluate the performance of a re-identification model, the dataset must represent commonly encountered confounding factors such as viewpoint and illumination variation. Hence, for each person, the images were acquired at two different times of the day (morning and afternoon), and for each of these two acquisitions, the viewpoints, and location of the camera were different. 

\begin{figure}[h]%
\centering
\includegraphics[width=8cm, natwidth =560, natheight = 420]{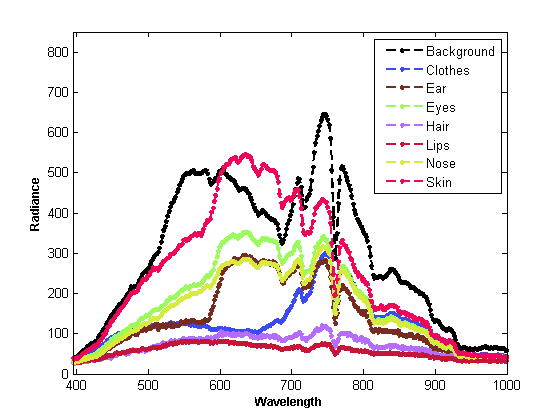} 
\includegraphics[width=8cm, natwidth =560, natheight = 420]{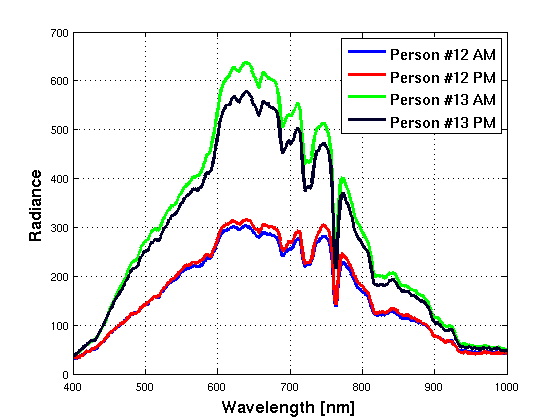} 
\caption{\emph{Top}: Mean spectral signatures for various material types in one of the images, including background, clothes, skin (from face and hands), and spectra from other parts of the head; \emph{Bottom}: Mean spectra of ``skin'' from people with IDs 12 and 13 in the gallery set (AM) and the probe set (PM).}
\label{fig:spectra}
\end{figure}

This data collection was carried out in two batches --- in the series of 15 hyperspectral images acquired in the morning, the camera orientation was fixed relative to the background in the scene, and each person was asked to arbitrarily stand somewhere in the scene and assume a natural (and arbitrary pose). Again, in the afternoon, another series of 15 images were acquired with the sample 15 people, but at a different location, and with a different orientation of the camera.  For the purpose of quantifying the potential of hyperspectral images for person re-identification, we treat the morning images from the 15 people as the gallery set, and the afternoon images from those people as the probe set. Also, in addition to the camera having two different orientations and locations in the morning and afternoon, each person was asked to stand arbitrarily in the scene in order to get arbitrary viewpoints. Due to the temporal difference of the captured data, variations in physiological conditions are also represented in this dataset.
Fig. \ref{fig:spectra} depicts spectral signatures from the background objects and various parts of the head from one of the $15$ people. It also depicts spectral signatures corresponding to skin pixels (from the face and hands) for two people (Person 12 and Person 13) in the gallery (AM) and probe (PM) sets. This suggests that hyperspectral data can effectively characterize variability between people when using pixels over skin objects.

\section{Spectral Angle based Person Re-identification}
\label{sec:method}

In this paper, we focus on the utility of spectral information from skin pixels for person re-identification.  Our approach consists of the following steps --- we manually annotated small patches of ``skin'' pixels (on the face and hands) for each person in the gallery and probe sets. We carried out superpixel segmentation to match mean spectra from skin superpixels to avoid pixel mixing that would result from rectangular windows that are traditionally employed. By averaging spectral signatures over superpixels, we stabilize spectral response from local variability and noise, while at the same time avoiding inadvertent mixing that would have resulted had we chosen rectangular windows. Specifically, we utilized the entropy rate superpixel algorithm which ensures compact and homogenous clusters of similar sizes \cite{ERSuperPixelLiu2011,priyasuperpixels2015}. We took superpixels that intersected with our ``ground-truth'' skin patches in each image and used those to match each probe image with the gallery.  A spectral angle distance was used as the similarity metric for the re-identification task. If $S_i$ and $S_j$ represent superpixels in the gallery and probe set respectively, then a spectral angle distance $d_{sa} (S_i,S_j)$ between the two superpixels can be defined as :

\begin{equation}\label{eq:SAD}
{d_{sa} (S_i,S_j)} = {\frac{{{\hat{{\mathbf{x}}}}_{S_i}^T}{{\hat{{\mathbf{x}}}}_{S_j}}}{{\parallel {{\hat{{\mathbf{x}}}}_{S_i}} \parallel} {\parallel {{\hat{{\mathbf{x}}}}_{S_j}} \parallel}}}
\end{equation}

\noindent where, ${\hat{{\mathbf{x}}}}_{S_i}$ is the mean spectral signature of region ${S_i}$. In addition to facilitating a computationally simple comparison, using spectral angle as a similarity metric with spectral signature features provides us with an added advantage in that the metric exhibits illumination invariance \cite{CP2015_ADA_JSTSP,keshava2004distance,adler2001shadow}. 
Average spectral angle distances between skin superpixels from the probe image and every image in the gallery set are used to re-identify the probe set. Note that the superpixels that result from the entropy rate superpixel algorithm modified to use spectral angle based similarity are highly effective at oversegmenting the image into uniform sized clusters while preserving borders, and are hence appropriate for spectral matching via the spectral angle distance. In a practical implementation, matching skin superpixels between probe and gallery sets can be easily accomplished by first utilizing a face detection algorithm to identify the skin superpixels in all the images.

%

\begin{figure}
   \centering
 \includegraphics[height = 7cm]{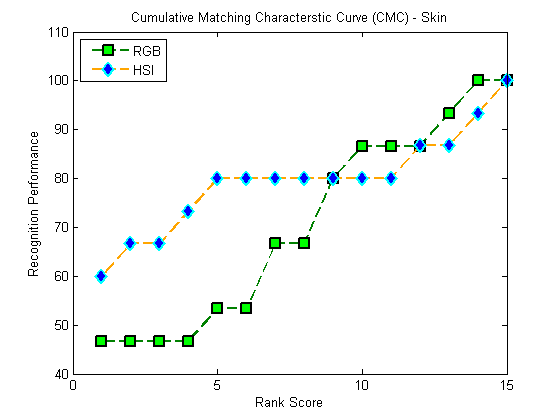}
\caption{CMC curves}\label{fig:CMC}				   
\end{figure}

For this closed set re-identification problem, we report performance via the cumulative matching characteristic (CMC) curve. 
The CMC curve represents the expectation of finding the correct match in the top \textit{{r}} matches. In other words, a rank-\textit{{r}} recognition rate shows the percentage of the test images that are correctly recognized from the top \textit{{r}} matches in the gallery set. The rank-1 value on this curve indicates the true identification performance, while the rank-$N$ score ($N$ being the number of images in the gallery) will be $100\%$ for closed-set re-identification, with the curve monotonically increasing from 1 through $N$. Comparison between hyperspectral and RGB (color) images based on this approach is reported in the CMC curve in Fig.~\ref{fig:CMC}. We observe that hyperspectral images not only provides a superior rank-1 performance compared to when using (RGB) color images, the CMC curve for hyperspectral images increases much faster as a function of the rank, with hyperspectral data providing as much as $20\%$ better identification performance.

\section{Conclusions, Caveats and Future Work}
We conclude from the data and the results that hyperspectral images have potential to enhance re-identification performance in multi-camera surveillance systems --- the preliminary dataset and resulting CMC curves suggest the ``value'' added by considering the narrow-band spectral information for re-identification as opposed to 3-channel color images. By being able to distinguish subtle spectral variations between people (e.g. via the spectral signatures of their skin), it may also enable long-duration re-identification wherein significant time may have elapsed between the person reappearing in the collective field of view of the system. We acknowledge an important caveat with results presented here --- the sample size (30 images total in gallery and probe sets) is small --- overall performance will drop when the number of people are added to the re-identification problem. However, we expect that spectral signature as a descriptor to characterize skin pixels effectively is likely to result in superior performance compared to traditional (RGB) color systems. Finally, we note that the approach to re-identification can be enhanced by complementing spectral signatures with other descriptors (e.g. spatial information). In ongoing work, we are expanding the gallery and probe sets, and are exploring strategies to fuse information provided from spectral signatures with currently established color image based feature descriptors for effective re-identification.

\balance

\bibliographystyle{IEEEbib}
\bibliography{strings,refs,TanuThesis,/users/sprasad/Dropbox/Research/BibFileSaurabh}

\begin{thebibliography}{10}

\bibitem{vyas2013estimating}
Saurabh Vyas, Amit Banerjee, and Philippe Burlina,
\newblock ``Estimating physiological skin parameters from hyperspectral
  signatures,''
\newblock {\em Journal of biomedical optics}, vol. 18, no. 5, pp.
  057008--057008, 2013.

\bibitem{li2012compressive}
Chengbo Li, Ting Sun, Kevin~F Kelly, and Yin Zhang,
\newblock ``A compressive sensing and unmixing scheme for hyperspectral data
  processing,''
\newblock {\em Image Processing, IEEE Transactions on}, vol. 21, no. 3, pp.
  1200--1210, 2012.

\bibitem{chen2011denoising}
G.~Chen and S.~Qian,
\newblock ``Denoising of hyperspectral imagery using principal component
  analysis and wavelet shrinkage,''
\newblock {\em IEEE Transactions on Geoscience and Remote Sensing}, vol. 49,
  no. 3, pp. 973--980, 2011.

\bibitem{NG2011}
N.~Gupta,
\newblock ``Development of staring hyperspectral imagers,''
\newblock in {\em Applied Imagery Pattern Recognition Workshop (AIPR), 2011
  IEEE}, 2011, pp. 1--8.

\bibitem{prasad2014SMoG}
S~Prasad, M.~Cui, W.~Li, , and J.~Fowler,
\newblock ``Segmented mixture of gaussian classification for robust sub-pixel
  hyperspectral {ATR},''
\newblock {\em IEEE Geoscience and Remote Sensing Letters}, vol. 11, no. 1, pp.
  138--142, January 2014.

\bibitem{PLFB2012}
S.~Prasad, W.~Li, J.~E. Fowler, and L.~M. Bruce,
\newblock ``Information fusion in the redundant-wavelet-transform domain for
  noise-robust hyperspectral classification,''
\newblock {\em Geoscience and Remote Sensing, IEEE Transactions on}, vol. 50,
  no. 99, pp. 3474--3486, September 2012.

\bibitem{KPB2010Derivatives}
H.R. Kalluri, S.~Prasad, and L.M. Bruce,
\newblock ``Decision-level fusion of spectral reflectance and derivative
  information for robust hyperspectral land cover classification,''
\newblock {\em Geoscience and Remote Sensing, IEEE Transactions on}, vol. 48,
  no. 11, pp. 4047 --4058, nov. 2010.

\bibitem{prasad2009information}
S.~Prasad and L.M. Bruce,
\newblock ``Information fusion in kernel-induced spaces for robust subpixel
  hyperspectral atr,''
\newblock vol. 6, no. 3, pp. 572--576, 2009.

\bibitem{YP2015}
Y.~Zhang and S.~Prasad,
\newblock ``Locality preserving composite kernel feature extraction for
  multi-source geospatial image analysis,''
\newblock {\em IEEE Journal of Selected Topics in Applied Earth Observations
  and Remote Sensing}, vol. 8, no. 3, pp. 1385--1392, March 2015.

\bibitem{PC2013Asilomar_Sparse}
S.~Prasad and M.~Cui,
\newblock ``Sparse representations for classification of high dimensional
  multi-sensor geospatial data,''
\newblock in {\em Proceedings of the 2013 Asilomar Conference on Signals,
  Systems and Computers.}, November 2013, pp. 811--815.

\bibitem{li2011multi}
W.~Li, S.~Prasad, J.E. Fowler, and L.M. Bruce,
\newblock ``A multi-modal pattern classification framework for hyperspectral
  image analysis,''
\newblock in {\em Hyperspectral Image and Signal Processing: Evolution in
  Remote Sensing (WHISPERS), 2011 3rd Workshop on}. IEEE, 2011, pp. 1--4.

\bibitem{di2011active}
W.~Di and M.M. Crawford,
\newblock ``Active learning via multi-view and local proximity
  co-regularization for hyperspectral image classification,''
\newblock {\em IEEE Journal of Selected Topics in Signal Processing}, vol. 5,
  no. 3, pp. 618--628, 2011.

\bibitem{PB2008a}
Saurabh Prasad and Lori~Mann Bruce,
\newblock ``Decision fusion with confidence-based weight assignment for
  hyperspectral target recognition,''
\newblock vol. 46, no. 5, pp. 1448--1456, May 2008.

\bibitem{gheissari2006person}
Niloofar Gheissari, Thomas~B Sebastian, and Richard Hartley,
\newblock ``Person reidentification using spatiotemporal appearance,''
\newblock in {\em IEEE Conference on Computer Vision and Pattern Recognition
  (CVPR)}. IEEE, 2006, vol.~2, pp. 1528--1535.

\bibitem{farenzena2010person}
Michela Farenzena, Loris Bazzani, Alessandro Perina, Vittorio Murino, and Marco
  Cristani,
\newblock ``Person re-identification by symmetry-driven accumulation of local
  features,''
\newblock in {\em IEEE Conference on Computer Vision and Pattern Recognition
  (CVPR)}. IEEE, 2010, pp. 2360--2367.

\bibitem{zheng2011person}
Wei-Shi Zheng, Shaogang Gong, and Tao Xiang,
\newblock ``Person re-identification by probabilistic relative distance
  comparison,''
\newblock in {\em IEEE Conference on Computer Vision and Pattern Recognition
  (CVPR)}. IEEE, 2011, pp. 649--656.

\bibitem{dantcheva2011frontal}
Antitza Dantcheva and J-L Dugelay,
\newblock ``Frontal-to-side face re-identification based on hair, skin and
  clothes patches,''
\newblock in {\em Advanced Video and Signal-Based Surveillance (AVSS), 2011 8th
  IEEE International Conference on}. IEEE, 2011, pp. 309--313.

\bibitem{ERSuperPixelLiu2011}
M.~Y. Liu, O.~Tuzel, S.~Ramalingam, and R.~Chellappa,
\newblock ``Entropy rate superpixel segmentation,''
\newblock in {\em Computer Vision and Pattern Recognition (CVPR), 2011 IEEE
  Conference on}, June 2011, pp. 2097--2104.

\bibitem{priyasuperpixels2015}
Tanu Priya, Saurabh Prasad, and Hao Wu,
\newblock ``Superpixels for spatially reinforced bayesian classification of
  hyperspectral images,''
\newblock {\em IEEE Geoscience and Remote Sensing Letters}, vol. 12, no. 5, pp.
  1071--1075, 2015.

\bibitem{CP2015_ADA_JSTSP}
Minshan Cui and Saurabh Prasad,
\newblock ``Angular discriminant analysis for hyperspectral image
  classification,''
\newblock {\em Selected Topics in Signal Processing, IEEE Journal of}, vol. 9,
  no. 6, pp. 1003--1015, 2015.

\bibitem{keshava2004distance}
Nirmal Keshava,
\newblock ``Distance metrics and band selection in hyperspectral processing
  with applications to material identification and spectral libraries,''
\newblock {\em IEEE Transactions on Geoscience and Remote Sensing (TGRS)}, vol.
  42, no. 7, pp. 1552--1565, 2004.

\bibitem{adler2001shadow}
Steven~M Adler-Golden, Robert~Y Levine, Michael~W Matthew, Steven~C
  Richtsmeier, Lawrence~S Bernstein, John~H Gruninger, Gerald~W Felde,
  Michael~L Hoke, Gail~P Anderson, et~al.,
\newblock ``Shadow-insensitive material detection/classification with
  atmospherically corrected hyperspectral imagery,''
\newblock in {\em Aerospace/Defense Sensing, Simulation, and Controls}.
  International Society for Optics and Photonics, 2001, pp. 460--469.

\end{thebibliography}

\end{document}